\documentclass[a4paper,twoside]{article}
\usepackage[top=33mm,bottom=42mm,left=26mm,right=26mm]{geometry}

\usepackage{epsfig}
\usepackage{paralist}
\usepackage{subcaption}
\usepackage{calc}
\usepackage{amssymb}
\usepackage{amstext}
\usepackage{amsmath}
\usepackage{amsthm}
\usepackage{multicol}
\usepackage{pslatex}
\usepackage{apalike}
\usepackage{tikz,ifthen}
\usepackage{url}
\usepackage{etoolbox}
\usepackage{float}
\usepackage{listings}
\usepackage{SCITEPRESS}     

\usetikzlibrary{arrows,trees,backgrounds,automata,shapes,decorations,plotmarks,fit,calc,positioning,shadows,chains,
  patterns,decorations.pathreplacing}
\tikzstyle{block}=[rectangle, draw, thin, inner sep=3pt, text centered, drop shadow, fill=orange!20!yellow!20]
\tikzstyle{pre}=[<-,shorten <=1pt,>=stealth']
\tikzstyle{post}=[->,shorten >=1pt,>=stealth']
\tikzstyle{bi}=[<->,shorten >=1pt,shorten <=1pt,>=stealth']
\tikzstyle{every initial by arrow}=[initial text={},initial distance=1em,post]
\tikzstyle{every state}=[minimum size=0.4cm,drop shadow,fill=orange!20!yellow!20]
\tikzstyle{transition}= [post,shorten >=1pt,node distance=2cm, inner sep=2pt,bend angle=20]
\usetikzlibrary{calc}

\usepackage{tikz,ifthen,pgfplots}
\pgfplotsset{compat=1.17}
\usetikzlibrary{arrows,trees,backgrounds,automata,shapes,decorations,plotmarks,fit,calc,positioning,shadows,chains}
\tikzstyle{every pin edge}=[<-,shorten <=1pt]
\tikzstyle{neuron}=[circle,fill=black!25,minimum size=17pt,inner sep=0pt]
\tikzstyle{input neuron}=[neuron, fill=green!50]
\tikzstyle{output neuron}=[neuron, fill=red!50]
\tikzstyle{hidden neuron}=[neuron, fill=blue!50]
\tikzstyle{annot} = [text width=6em, text centered]
\usetikzlibrary{calc}

\DeclareFixedFont{\ttb}{T1}{txtt}{bx}{n}{7.5} 
\DeclareFixedFont{\ttm}{T1}{txtt}{m}{n}{7.5}  

\definecolor{deepblue}{rgb}{0,0,0.5}
\definecolor{deepred}{rgb}{0.6,0,0}
\definecolor{deepgreen}{rgb}{0,0.5,0}

\renewcommand\j[1]{\textsc{#1}}
\newcommand{\originalAgent}{$\mathcal{A_O}$}
\newcommand{\enhancedAgent}{$\mathcal{A_E}$}

\begin{document}
\title{Scenario-Assisted Deep Reinforcement Learning}

\author{\authorname{Raz Yerushalmi\sup{1}, Guy Amir\sup{2}\orcidAuthor{0000-0002-7951-7795}, Achiya Elyasaf\sup{3}\orcidAuthor{0000-0002-4009-5353}, 
David Harel\sup{1}, Guy Katz\sup{2} and Assaf Marron\sup{1}\orcidAuthor{0000-0001-5904-5105}}
\affiliation{\sup{1}Department of Computer Science and Applied Mathematics, Weizmann Institute of Science, Rehovot 76100, Israel}
\affiliation{\sup{2}School of Computer Science and Engineering, The Hebrew University of Jerusalem, Givat Ram, Jerusalem 91904, Israel}
\affiliation{\sup{3}Software and Information Systems Engineering, Ben-Gurion University of the Negev, Beer-Sheva 8410501, Israel}
\email{\{raz.yerushalmi, david.harel, assaf.marron\}@weizmann.ac.il, \{guyam, guykatz\}@cs.huji.ac.il, achiya@bgu.ac.il}
}


\abstract{Deep reinforcement learning has proven remarkably useful in
  training agents from unstructured data. However, the opacity of the
  produced agents makes it difficult to ensure that they adhere to
  various requirements posed by human engineers.  In this work-in-progress report, we propose a
  technique for enhancing the reinforcement learning training process
  (specifically, its reward calculation), in a way that allows human
  engineers to directly contribute their expert knowledge, making the
  agent under training more likely to comply with various relevant
  constraints. Moreover, our proposed approach allows formulating
  these constraints using advanced model engineering techniques, such
  as scenario-based modeling. This mix of black-box learning-based
  tools with classical modeling approaches could produce systems that are
  effective and efficient, but are also more transparent and
  maintainable.  We evaluated our technique using a case-study from the
  domain of internet congestion control, obtaining promising results.}

\onecolumn \maketitle \normalsize \setcounter{footnote}{0} \vfill

\section{\uppercase{Introduction}}
\label{sec:introduction}

\emph{Deep neural networks} (\emph{DNNs}) have proven highly
successful in addressing hard-to-specify cognitive tasks. \emph{Deep
  reinforcement learning} (\emph{DRL}) is a particular method for
producing DNNs, which is applicable in cases where the training data
is unstructured --- e.g., in games~\cite{Ye_Fu_Yang_Huang_2020}, in
autonomous driving~\cite{Kiran9351818_2021}, smart city
communications~\cite{Xia91308782021},
manufacturing~\cite{LI2022104957}, chat
bots~\cite{MOHAMADSUHAILI2021115461}, context-aware systems~\cite{elyasaf2021context}, and many others. As this trend
continues, it is likely that DRL will gain a foothold in many systems of
critical importance.

Despite the success of DRL, and in particular its generally superior
performance to that of hand-crafted code in many kinds of applications, various problematic aspects
of this paradigm have begun to emerge. One issue is that DRL agents
are typically trained on some distribution of inputs (described, e.g.,
using a Markov Decision Process), but this distribution might
differ from the distribution that the agent encounters after
deployment~\cite{ElKaKaSc21}. Another issue is various
vulnerabilities that exist in many DNNs, and in particular in DRL
agents~\cite{SzZaSuBrErGoFe13}, such as sensitivity to adversarial perturbations.  
When such issues are discovered, the
DRL agent often needs to be modified or enhanced; but unfortunately,
such routine actions are known to be extremely difficult for DRL
agents, which are largely considered ``black
boxes''~\cite{FoBeCu16}. Specifically, their underlying DNNs are opaque
to the human eye, making them hard to interpret; and the DRL training
process itself is computationally expensive and time-consuming, making
it infeasible to retrain the agent whenever circumstances, or
requirements, change.
Much research is being conducted on DNN interpretability and
explainability~\cite{RiSiGu16,SaWiMu18} as well as on using formal
methods to facilitate reasoning about DRL agents~\cite{KaBaKaSc19},
but these efforts are still nascent, and typically suffer from limited
scalability.

In this paper, we advocate a direction of work for addressing this important
gap, by integrating \emph{modeling techniques} into the DRL training
process. The idea is to leverage the strengths of classical
specification approaches, which are normally applied in procedural or
rule-based modeling, and carry these advantages over to the training
process of DRL agents. More specifically, we propose to sometimes
override the computation of the DRL agent's \emph{reward function}~\cite{rl-book-new}, which is then reflected in the \emph{return} that the agent is being trained to maximize. 
By creating a connection that will allow
modelers to formulate a specification in their modeling formalism of
choice, and then have this specification affect the computed reward so
that it reflects how well the specification is satisfied, we seek to
generate a DRL agent that better conforms to the system's
requirements.
 
Our proposed approach is general, in the sense that numerous modeling
formalisms could be integrated into the DRL process. As a
proof-of-concept, and for evaluation purposes, we focus here on a
particular brand of modeling schemes, collectively referred to as
\emph{scenario-based modeling}
(\emph{SBM})~\cite{DaHa01,HaMaWe12ACM}. In SBM, a modeler creates
small, stand-alone scenarios, each reflecting a certain desirable or
undesirable behavior of the system under development.  These scenarios
are fully executable, and when interwoven together bring about an executable model of
the desired global system behavior. A key feature of SBM is the ability
of each scenario to specify \emph{forbidden} behavior, which the
system as a whole should avoid. SBM has been shown to be quite
effective in modeling systems from varied domains, such as
web-servers~\cite{HaKa14}, cache coherence protocols~\cite{KaBaHa15},
games~\cite{HaLaMaWe11BPMC}, production
control~\cite{harelKuglerWeiss2005LSCproductionControl}, biological
systems~\cite{kugler2008LSCcElegansDev},
transportation~\cite{GrGrKaMa16} and others.

During DRL training, the agent may be regarded as a reactive system:
it receives an input from the environment, reacts, observes the
computed reward that its actions have produced, and optionally adjusts
its behavior for the future.  In order to integrate SBM and DRL, we
suggest to execute the scenario-based model in parallel to the DRL
agent's training.  Then, whenever the agent performs an action that
the SBM model forbids, we propose to reflect this by penalizing the
agent, through its reward values.  We argue that this process would
increase the likelihood that the agent learns, in addition to its
original goals, the constraints expressed through the scenario-based
model.

As a case study, we chose to focus on the Aurora
system~\cite{pmlr-v97-jay19a}, which is a DRL-based Internet
congestion control algorithm. The algorithm is deployed at the sender
node of a communication system, and controls the sending rate of that
node, with the goal of optimizing the communication system's
throughput. As part of our evaluation, we demonstrate how a
scenario-based specification can be used to influence the training of
the Aurora DRL agent, in order to improve its \emph{fairness} by
preventing it from repeatedly increasing its sending rate at the
possible expense of other senders on the same link. Our experiments
show that an agent trained this way is far less likely to exhibit the
unwanted behavior, when compared to an agent trained by DRL alone, thus
highlighting the potential of our approach.

The rest of the paper is organized as follows. In Section
\ref{background} we provide background on scenario-based modeling and
deep reinforcement learning. In Sections~\ref{integratingSBM_DRL}
and~\ref{technicaldescription} we describe the integration between SBM
and DRL, first conceptually and then technically.  In
Section~\ref{casestudy} we describe our case study. We follow with a
discussion of related work in Section~\ref{relatedwork} and conclude
in Section~\ref{conclusion}.

\section{\uppercase{Background}}\label{background}

\subsection{Scenario-Based Modeling} \label{SBPintro}
Scenario-based modeling
(SBM)~\cite{DaHa01,harelMarelly2003comePlay,HaMaWe12ACM} is a modeling
paradigm, designed to facilitate the development of reactive systems
from components that are aligned with the way humans perceive and
describe system behavior.  The focus of SBM is on inter-object,
system-wide behaviors, thus differing from the more conventional,
object-centric paradigms. In SBM, a system is comprised of components
called \emph{scenario objects}, each of which describes a single
desired or undesired behavior of the system. This behavior is
formalized as a sequence of events.
A scenario-based model is fully executable: when it runs, all its
scenario objects are composed and run in parallel, in a synchronized
fashion, resulting in cohesive system behavior.  The resulting model
thus complies with the requirements and constraints of each of the
participating scenario
objects~\cite{harelMarelly2003comePlay,HaMaWe12ACM}.

More concretely, each scenario object in a scenario-based model can be
regarded as a transition system, whose states are referred to as
\emph{synchronization points}. The scenario object transitions between
synchronization points according to the triggering of \emph{events} by
a global \emph{event selection mechanism}. At each synchronization
point, the scenario object affects the triggering of the next event by
declaring events that it \emph{requests} and events that it
\emph{blocks}. These declarations encode, respectively, desirable and
forbidden actions, as seen from the perspective of that particular
scenario object. Scenario objects can also declare events that they
passively \emph{wait-for}, thus asking to be notified when these
occur. After making its event declarations, the scenario object is
suspended until an event that it requested or waited-for is triggered,
at which point the scenario resumes and may transition to another
synchronization point. 

At execution time, all scenario objects are run
in parallel, until they all reach a synchronization point. Then, the
declarations of all requested and blocked events are collected by the
event selection mechanism, which first selects, and then triggers one of the
events that is requested by at least one scenario object and is
blocked by none.

Fig.~\ref{fig:watertap} (borrowed from~\cite{HaKaMaWe12}) depicts a
scenario-based model of a simple system for controlling the
temperature and fluid level in a water tank. Each scenario object is
depicted as a transition system, in which the nodes represent
synchronization points. The transition edges are associated with the
requested or waited-for events in the preceding node. The scenarios
\j{AddHotWater} and \j{AddColdWater} repeatedly wait for \j{WaterLow}
events and then request three times the event \j{AddHot} or
\j{AddCold}, respectively.  Since, by default, these six events may be
triggered in any order, a new scenario \j{Stability} is introduced,
with the intent of keeping the temperature more stable. It enforces
the interleaving of \j{AddHot} and \j{AddCold} events by alternately
blocking them. The resulting execution trace is depicted in the event
log.
    
 \newcommand{\request}{{\color{blue}request}}
\newcommand{\waitfor}{{\color{green!50!black}wait for}}
\newcommand{\blocking}{{\color{red}blocking}}
\begin{figure}[hp]
  \centering
  \scalebox{0.65} {

    \tikzstyle{box}=[draw,  text width=2cm,text centered,inner sep=3]
    \tikzstyle{set}=[text centered, text width = 10em]

    \begin{tikzpicture}[thick,auto,>=latex',line/.style ={draw, thick, -latex', shorten >=0pt}]

      \matrix(bts) [row sep=0.3cm,column sep=2cm]  {

        \node (box1)  [box] {\waitfor{} \j{WaterLow}}; \\
        \node (box2)  [box] {\request\ \j{AddHot}}; \\
        \node (box3)  [box] {\request\ \j{AddHot}}; \\
        \node (box4)  [box] {\request\ \j{AddHot}}; \\
      };

      \node (title) [above=0.1cm of bts,box,draw=none] at ($(bts) + (-0.25cm,2.31cm)$)
      {\j{AddHotWater}};

      \begin{scope}[every path/.style=line]
        \path (box1)   -- (box2);
        \path (box2)   -- (box3);
        \path (box3)   -- (box4);
        \path (box4.east)   -- +(.25,0) |- (box1);
      \end{scope}

      \matrix(bts2) [right=.25cm of bts, row sep=0.3cm,column sep=2cm] {
        \node (box1)  [box] {\waitfor{} \j{WaterLow}}; \\
        \node (box2)  [box] {\request\ \j{AddCold}}; \\
        \node (box3)  [box] {\request\ \j{AddCold}}; \\
        \node (box4)  [box] {\request\ \j{AddCold}}; \\
      };

      \node (title) [above=0.1cm of bts2,box,draw=none] at ($(bts2) + (-0.25cm,2.31cm)$)
      {\j{AddColdWater}};

      \begin{scope}[every path/.style=line]
        \path (box1)   -- (box2);
        \path (box2)   -- (box3);
        \path (box3)   -- (box4);
        \path (box4.east)   -- +(.25,0) |- (box1);
      \end{scope}

      \matrix(bts3) [right=.25cm of bts2, row sep=0.3cm,column sep=2cm] {
        \node (box1)  [box] {\waitfor{}  \j{AddHot} while  \blocking\ \j{AddCold}}; \\
        \node (box2)  [box] {\waitfor{}  \j{AddCold} while \blocking\ \j{AddHot}}; \\
      };

      \node (title) at (title-|bts3) [box,draw=none] {\j{Stability}};

      \begin{scope}[every path/.style=line]
        \path (box1)   -- (box2);
        \path (box2.east)   -- +(.25,0) |- (box1);
      \end{scope}

      \node (log)  [right=.3cm of bts3,box,text width=2cm,fill=yellow!20] {
        $\cdots$ \\
        \j{WaterLow} \\
        \j{AddHot}  \\
        \j{AddCold} \\
        \j{AddHot}  \\
        \j{AddCold} \\
        \j{AddHot}  \\
        \j{AddCold} \\
        $\cdots$ \\
      };

      \node (title2) at (title-|log)
      [box,draw=none] {\j{Event Log}};
    \end{tikzpicture}
  }
  \caption{(Borrowed from~\cite{HaKaMaWe12}) A scenario-based model for controlling water temperature and
      level. }
  \label{fig:watertap}
\end{figure}
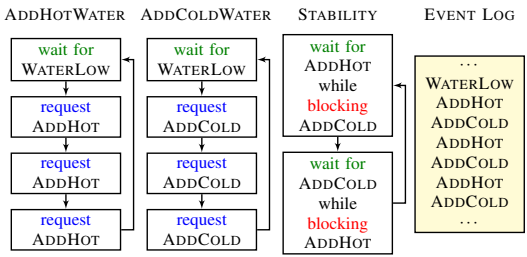

Selecting the next event to be triggered, from among all events that are
requested and not blocked, can be customized to fit the system at
hand. Common policies include arbitrary selection, randomized
selection, a selection based on predefined priorities, or a selection
based on look-ahead for achieving certain
outcomes~\cite{kugler2002smart}.

For our present purposes, it is convenient to think of scenario
objects in terms of transition systems. In practice, SBM is supported
in a variety of textual and visual frameworks. Notable examples
include the language of \emph{live sequence charts} (LSC), where SBM
concepts are applied to produce modal sequence
diagrams~\cite{DaHa01,harelMarelly2003comePlay}; implementations in
various high level languages, such as Java, C, C++, Erlang,
JavaScript, and Python (see, e.g.,~\cite{HaMaWe12ACM}); and various
domain specific languages~\cite{GrGrKaMa16} and
extensions~\cite{HaKaMaSaWe20,KaMaSaWe19}.

A particularly useful trait of SBM is that the resulting models are
amenable to model checking, and facilitate compositional
verification~\cite{HaLaMaWe11BPMC,HaKaKaMaMiWe13,HaKaMaWe15,KaBaHa15,Ka13,HaKaLaMaWe15}. Thus,
it is often possible to apply formal verification to ensure that a
scenario-based model satisfies various criteria, either as a
stand-alone model or as a component within a larger system. Automated
analysis techniques can also be used to execute scenario-based models
in distributed
architectures~\cite{HaKaKaMaWeWi15,StGrGrHaKaMa18,StGrGrHaKaMa17,GrGrKaMaGlGuKo16,HaKaKa13},
to automatically repair these
models~\cite{HaKaMaWe14,HaKaMaWe12,Ka21}, and to augment them in
various ways, e.g., as part of the Wise Computing
initiative~\cite{HaKaMaMa18,MaArElGoKaLaMaShSzWeHa16,HaKaMaMa16,HaKaMaMa16b}.

For our work here, namely the injection of domain-specific knowledge
into the DRL training procedure, SBM is an attractive choice, as it is
formal, executable, and facilitates incremental
development~\cite{GoMaMe12Spaghetti,AlArGoHa14CognitiveLoad}.
Furthermore,
its natural alignment with how experts may describe the
specification of the system at hand helps in transparently highlighting important parts of the training procedure. Indeed,
using SBM to complement DRL was demonstrated in the past, although the focus
so far has been on \emph{guarding} an already-trained DRL agent,
rather than 
affecting what the agent actually learns~\cite{katz2020guarded,katzElyasaf2021DLSBP}.

\subsection{Deep Reinforcement Learning}
Deep reinforcement learning~\cite{rl-book-new} is a method for
automatically producing a decision-making agent, whose goal during
training is to achieve a high \emph{return} (according to some function)
through interactions with its environment.

\begin{figure}[htp]
	\centering
	\includegraphics[width=0.7\linewidth]{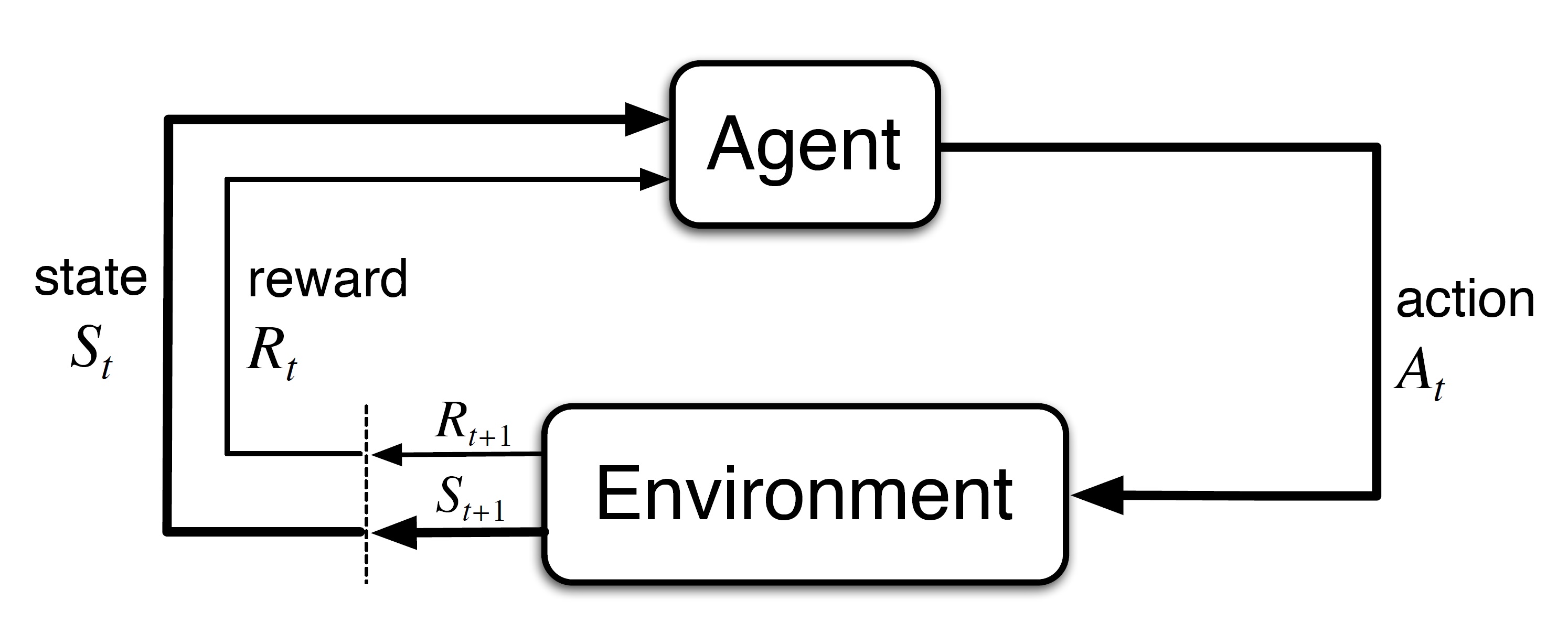}
	\caption{(Borrowed from~\cite{rl-book-new}) The
          agent-environment interaction in reinforcement learning. }
	\label{fig:agent-environment-interaction-mdp}
\end{figure}

Fig.~\ref{fig:agent-environment-interaction-mdp} depicts the basic DRL
learning cycle. 
The agent and its environment interact at discrete time steps
$t \in \{0,1,2,3, \ldots \}$. At each time step $t$, the agent
 observes the environment's state $s_{t}$, and selects its
action $a_{t}$ accordingly. In the subsequent time step ${t+1}$, and
as a result of its action $a_{t}$ at time ${t}$, the agent receives
its reward 
$R_{t}=R(s_{t}, a_{t})$, 
the environment moves to state $s_{t+1}$, and the
process repeats.
Through this interaction, the agent gradually learns a 
 policy function
$f \colon s_{t} \to a_{t}$ {}
that maximizes its \emph{return} $G_{t}$, the 
future cumulative discounted reward:
\[
  G_{t} = \sum_{k=0}^{\infty} \gamma^{k}R_{t+k+1}
\]
where $R_{t}$ is the reward at time $t$, and $\gamma$ is a \emph{discount rate} parameter, $0 \leq \gamma \leq 1$.

It is commonly accepted that specifying an appropriate reward function
$R_{t}$ (as function of the action $a_{t}$ in state $s_{t}$) 
is
crucial to the success of the  
DRL training process.  Consequently, this
topic has received significant
attention~\cite{Ng1999PolicyIU,zou2019reward,rl-book-new}.  As we
later explain, the approach that we advocate here is complementary to
this line of research: we propose to augment the reward function 
with constraints and specifications provided by domain experts.

\section{\uppercase{Integrating SBM into the Reward Function}}\label{integratingSBM_DRL}

Our proposed approach is to integrate a scenario-based model into the
DRL training loop, in order to instruct the agent being trained to
follow the constraints and specifications embodied in those
scenarios. Specifically, we 
create a one-to-one mapping between the DRL agent's possible actions and a dedicated subset of the events in the scenario-based model,
so that the scenario objects may react to the agent's actions.  

We execute the scenario-based model alongside the agent
under training, and, if at time step $t$ we denote 
the model's state $\tilde{s}_t$, the agent's reward function $R_t$
is computed as follows:
\begin{inparaenum}[(i)]
\item	at time step $t$, the agent selects an action $a_t$;
\item the environment reacts to $a_t$, transitions to a new state $s_{t+1}$, and computes a candidate reward value $\tilde{R_t}$;
\item the scenario-based model also receives $a_t$, and
  transitions to a new state $\tilde{s}_{t+1}$; 
\item if $a_t$ \emph{is} blocked in state $\tilde{s}_{t}$, the
  scenario-based model
\emph{penalizes}  the agent by 
decreasing the reward:
\begin{equation}
  R_t=
  \begin{cases}
    \alpha\cdot\tilde{R_t} - \Delta & ; \text{ if }
    \tilde{s}_t\xrightarrow{a_t}\tilde{s}_{t+1} \text{ is blocked}
    \\
    \tilde{R_t} & ; \text{ otherwise}  \\
  \end{cases}
      \label{eq:modifiedReward}
    \end{equation}
    for some constants $\alpha\in [-1, 1]$ and $\Delta\geq 0$; and
\item $R_t$ is returned to the agent as the reward value at this step.
\end{inparaenum}

\begin{figure}[H]
	\centering
	\includegraphics[width=0.8\linewidth]{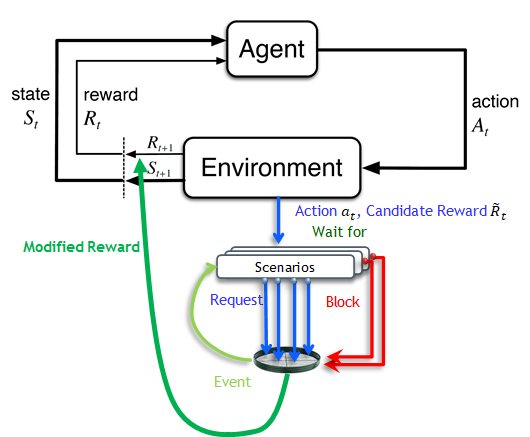}
	\caption{The agent-environment interaction in DRL integrated
          with SBM: At each time-step, the environment calculates a candidate 
          reward $\tilde{R_t}$ based on its state $s_{t}$, and the agent
          action $a_{t}$. The scenario-based model is executed in
          parallel, and may reduce $\tilde{R_{t}}$ if $a_{t}$ is a forbidden action in its state $\tilde{s}_{t}$.}
	\label{fig:agent-environment-SBP-intearction-in-reinforcement-leanring}
\end{figure}

The new training process is illustrated in
Fig.~\ref{fig:agent-environment-SBP-intearction-in-reinforcement-leanring}.
The motivation for these changes in the training process is to allow
the DRL agent to learn a policy that satisfies the requirements
encoded in the original reward function, while at the same time learning to
satisfy the specifications encoded as scenarios.  This is done without
changing the interface between the agent and its learning
environment. Observe that the scenario-based model is currently aware
of its own internal state and of the actions selected by the DRL
agent; but is unaware of the environment state $s_t$. Allowing the
scenario objects to view also the environment state, which may be
required in more complex systems, is left for future work.

Selecting an appropriate penalty policy, i.e. selecting the $\alpha$
and $\Delta$ constants, can of course have a significant impact on
the learned policy.  Consideration should be given to the balance
between allowing the agent to learn a policy that solves the original
problem, and encouraging it to follow the SBM specifications. 

Let us consider again the ``hot/cold'' example of Fig.~\ref{fig:watertap},
this time from a DRL perspective. Suppose we want to train a DRL
agent whose goal is to keep the water temperature steady, by mixing
hot and cold water. There are many strategies that would achieve this
goal; but suppose we also wish to introduce a constraint that the
agent should avoid two consecutive additions of hot water or of cold water.
We can achieve this by creating
a simple scenario-based model comprised of the \emph{Stability}
scenario (Fig.~\ref{fig:watertap}), and integrating
it into the training process. This scenario object would then penalize the DRL
agent whenever it performs the undesirable sequence of actions,
pushing it towards learning the desired policy.

\section{\uppercase{Proof-of-Concept Implementation}}
\label{technicaldescription}

We have created a simple proof-of-concept implementation of our approach,
by extending the AI-Gym Python framework for training DRL
agents \cite{brockman2016openai}. Specifically, we connected the AI-Gym
environment object (\emph{env})
to the BP-Py framework for specifying scenario-based models in
Python~\cite{bppycodebgu},
in a way that allows scenario objects to
affect the computation of the reward function  by the \emph{env}, as previously described.

The main connection point is AI-Gym \emph{env}'s \emph{step} method, which is
invoked in every iteration of the DRL training process, and which
eventually computes the reward value.  We altered the method so that
it communicates with the SBM core, to inform the scenario objects of the agent's
selected action, and in turn to receive instructions on how to modify the reward
value, if needed.

Recall that normally, a scenario-based model will, in each iteration:
\begin{inparaenum}[(i)]
\item trigger an event that is requested and not blocked; and
\item wake up all scenario objects that requested or waited-for the
  triggered event, allowing them to react by transitioning to a new
  state and updating their respective declarations of requested,
  blocked and waited-for events.
\end{inparaenum}
This high-level loop appears in
  Fig.~\ref{fig:sbp_mainloop}. The execution terminates once there are
  no enabled events, e.g., events that are requested and not blocked.

\begin{figure}[ht]
\begin{lstlisting}
# Main loop
while True:
    if noEnabledEvents():
        terminateExecution()

    event = selectEnabledEvent()
    advanceAllBThreads(event)
\end{lstlisting}
\caption{Pseudo code of the main execution loop in a regular scenario-based model. The execution will terminate once there are no enabled events.}
\label{fig:sbp_mainloop}
\end{figure}

In order to support integration with AI-Gym, and allow the
scenario-based model to execute in parallel to the training of the DRL
agent, we modified the execution scheme to
run in \emph{super steps}~\cite{kugler2002smart}: the SBM
model runs until it has no additional enabled events, and then,
instead of terminating, it waits for a new action event to be selected
by the DRL agent. Once such an event is triggered by the agent, it is
processed by the scenario objects (like any other triggered
event). However, if the event triggered by the agent happens to be
blocked by the scenarios, information is passed back to the AI-Gym 
\emph{env} to penalize the agent's reward value. The scenario objects
then carry out another super step, and the process repeats.
Fig.~\ref{fig:sbp_mainloop_ext} shows a pseudo code of the modified,
high-level loop.

\begin{figure}[ht]
\begin{lstlisting}
# Modified main loop, with super steps
while True:
    # Perform super step
    while haveEnabledEvents():
        event = selectEnabledEvent()
        advanceAllBThreads( event )

    # Handle an agent action
    action = waitForAgentAction()
    if isBlocked( action ):
        penalizeAgentReward()
    else:
        keepOriginalAgentReward()
    advanceAllBThreads( action )
\end{lstlisting}
\caption{Pseudo code of the main loop of the scenario-based execution  
  integrated with the DRL training. After each super step,
  the model waits for an action from the agent, and penalizes the
  agent if that action was blocked.}
\label{fig:sbp_mainloop_ext}
\end{figure}

\section{\uppercase{Case Study: The Aurora Congestion Controller}}
\label{casestudy}

As a case study, we chose to focus on the Aurora
system~\cite{pmlr-v97-jay19a} --- a DRL-based Internet
congestion control algorithm. The algorithm is deployed at the sender
node of a communication system, and controls the sending rate of that
node, with the goal of optimizing the communication system's
throughput. The selection of sending rate is based on various
parameters, such as previously observed throughput, the link's latency, and
the percent of previously lost packets. Aurora is intended to replace
earlier hand-crafted algorithms for obtaining similar goals, and was
shown to achieve excellent performance~\cite{pmlr-v97-jay19a}.

The authors of Aurora raised an interesting point regarding its
fairness, asking: \emph{Can our RL agent be trained to ``play well''
  with other protocols (TCP, PCC, BBR,
  Copa)?}~\cite{pmlr-v97-jay19a}. 
Indeed, in the case of Aurora, and
more generally in DRL, it is often hard to train the agent to comply
with various fairness properties.  In our case study, we set out to
add specific fairness constraints to the Aurora agent, using SBM.
Specifically, we attempted to avoid the situation where the algorithm 
becomes a
``bandwidth hog'' --- i.e., to train it not to increase its sending
rate continuously, thus providing other senders on the same link with a
fair share of the bandwidth.

\subsection{Evaluation Setup}

Using the BP-Py environment~\cite{bppycodebgu}, we created a simple
scenario-based model. This model, comprised of a single scenario, is
designed to penalize the Aurora agent for $k$ consecutive increases in
sending rate.  The SBM code for $k=3$ appears in
Fig.~\ref{fig:avoid3inraw}.

\begin{figure}[H]
\begin{lstlisting}
def SBP_avoid_k_in_a_row():
    k = 3
    counter = 0
    blockedEvList = []
    waitforEvList = [BEvent("IncreaseRate"),
                       BEvent("DecreaseRate"), 
                       BEvent("KeepRate")]
    while True:
        lastEv = yield{ waitFor:waitforEvList,
                        block:blockedEvList }
        if lastEv != None:
            if lastEv == BEvent("DecreaseRate") 
            or lastEv == BEvent("KeepRate"):  
                counter = 0
                blockedEvList = []
            else:
                if counter == k - 1:
                    #Blocking!
                    blockedEvList.append(
                      BEvent("IncreaseRate"))
                else:
                    counter += 1
\end{lstlisting}
\caption{The Python implementation of a scenario that blocks the \emph{IncreaseRate} event after $k-1$ consecutive occurrences.}
\label{fig:avoid3inraw}
\end{figure}

The scenario waits for three possible events that represent the three
different actions of the Aurora agent: \emph{IncreaseRate},
\emph{DecreaseRate}, and \emph{KeepRate}, representing the agent's
decision to increase the sending rate, decrease it or keep it steady, respectively. Whenever the \emph{IncreaseRate} event is triggered
$k-1$ times consecutively, the scenario will block it, until a
different action is selected by the agent.  Once the execution
environment detects that a requested agent action maps to a blocked
event, it will override the reward with a penalty, thus signaling to the
agent that it is not a desirable behavior.

For training the Aurora agent, we used the original training framework
provided in~\cite{pmlr-v97-jay19a}. We compared an agent trained using
the original framework, \originalAgent{}, to an agent trained using our
SBM-enhanced framework, \enhancedAgent{}. All Aurora-related
configurable parameters were identical between the two agents.

For computing the penalty when training \enhancedAgent{}, we
empirically selected the penalty function parameters (as described by
equation~\ref{eq:modifiedReward}) to be $\alpha = 0, \Delta =
-4.5$. These parameters allowed the agent to effectively learn both the
main goals and the additional constraints specified by the
scenario-based model. Values of $\Delta$ in the range $(-2, 0]$
proved to be too small (the agent failed to address the SBM
constraints), whereas values in the range $(-\infty,-10]$ resulted in
a disruption to the process of the agent learning its main goals.

\subsection{Evaluation Results}

\begin{figure}[H]
	\centering
	\includegraphics[width=1.0\linewidth]{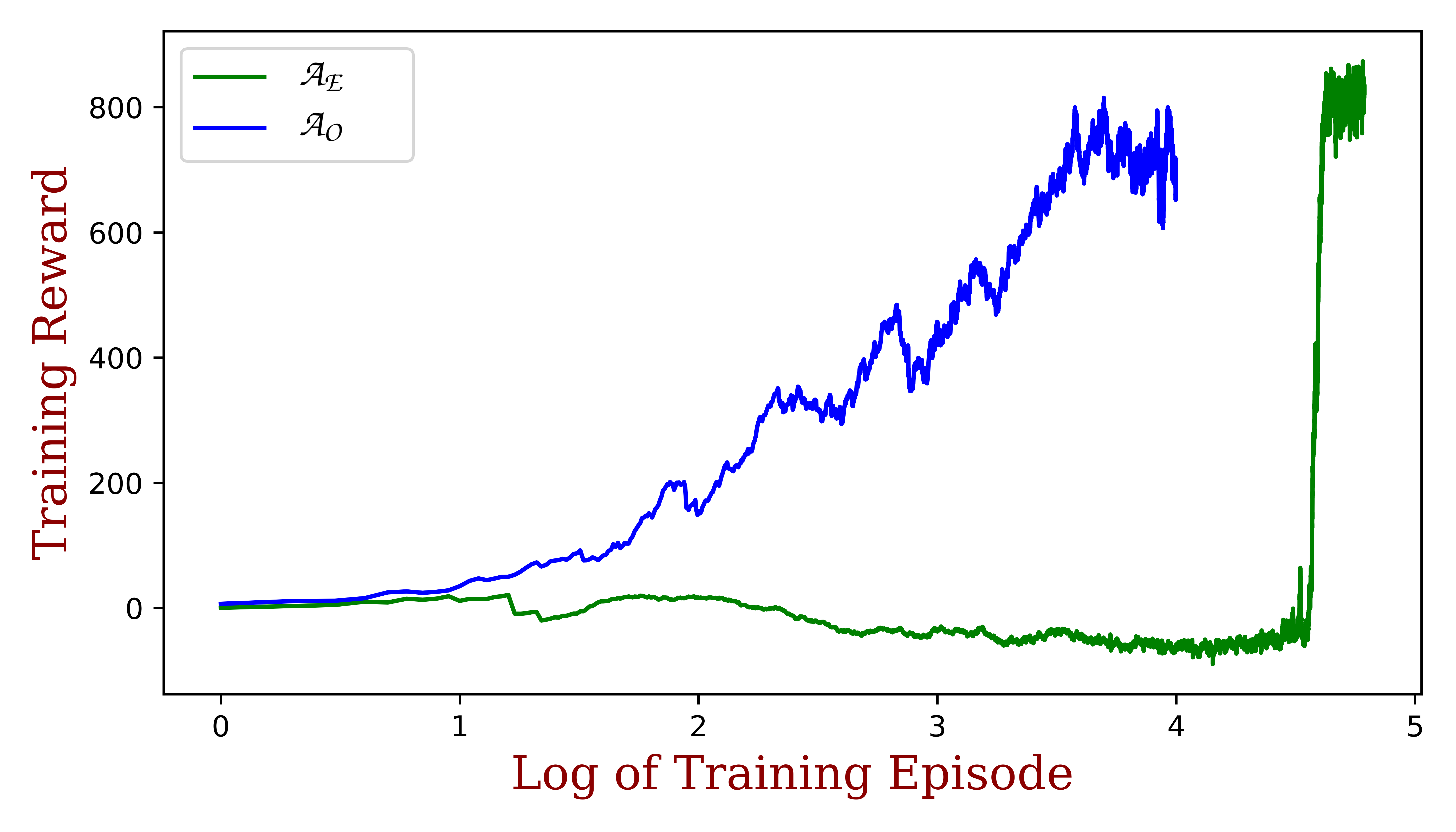}
	\caption{The average training reward obtained by \originalAgent{} and
          \enhancedAgent{}, as a function of time (log scaled). }
	\label{fig:alltogether_ewma_reward_logscale}
\end{figure}

We begin by comparing the two agents employing the evaluation metrics used
by~\cite{pmlr-v97-jay19a}.
Fig. \ref{fig:alltogether_ewma_reward_logscale} shows the training
performance of both \originalAgent{} and \enhancedAgent{} as a
function of the time-step (in log scale).

These results reveal a significant difference in the training times of
the two agents. Specifically, it takes  \enhancedAgent{}
significantly more epochs to learn an ``adequate'' policy, i.e. to
reach a similar reward level to that obtained by \originalAgent{}. We
observe that \enhancedAgent{} converges to a good policy after about
40,000 epochs (a little after 4.5 in the logarithmic time-step scale),
compared to about 3000 epochs of \originalAgent{} (around 3.5 in the
logarithmic time-step scale).

\begin{figure}[H]
  \centering
  \includegraphics[width=0.9\linewidth]{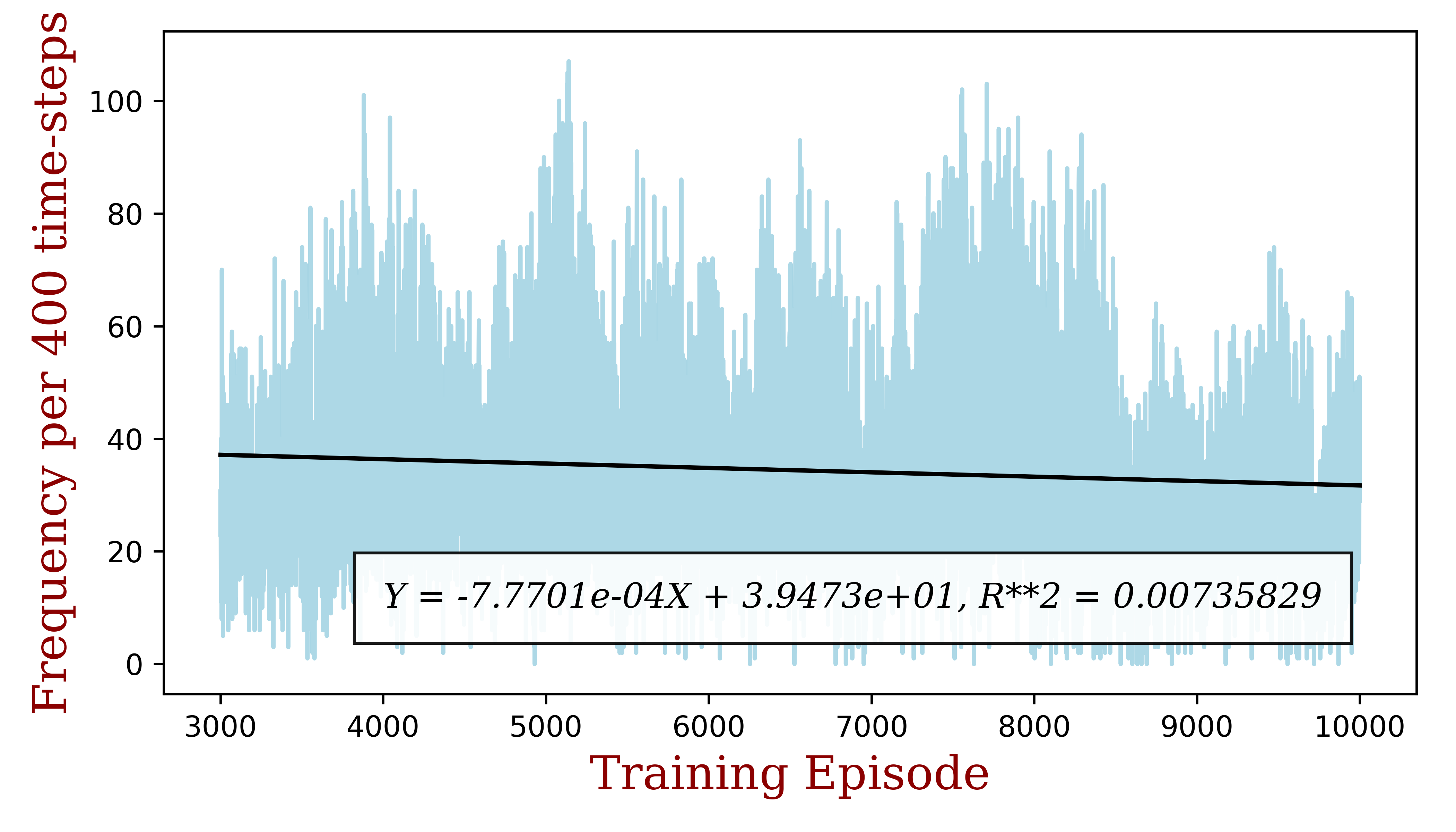}
  \caption{Each line shows the number of times \originalAgent{} chose
    to increase the sending rate three consecutive times or more,
    during the training process. Each training episode is comprised of
    400 time
    steps, and the agent performed on average 35-45 violations per episode, which translates to a frequency of about 9\%-11\%. The resulting linear regression line is $y = -0.00077701x + 39.47343569$.
    }
	\label{fig:Aurora_training_3some_fequency_plot}
\end{figure}

Next, we compare the frequency of the two agents choosing to increase
their sending rate three consecutive times or more (performing a
``violation''). For \originalAgent{}, the average frequency of such
violations is 9\%-11\%, as can be seen in
Fig.~\ref{fig:Aurora_training_3some_fequency_plot}.

\begin{figure}[H]
	\centering
	\includegraphics[width=0.9\linewidth]{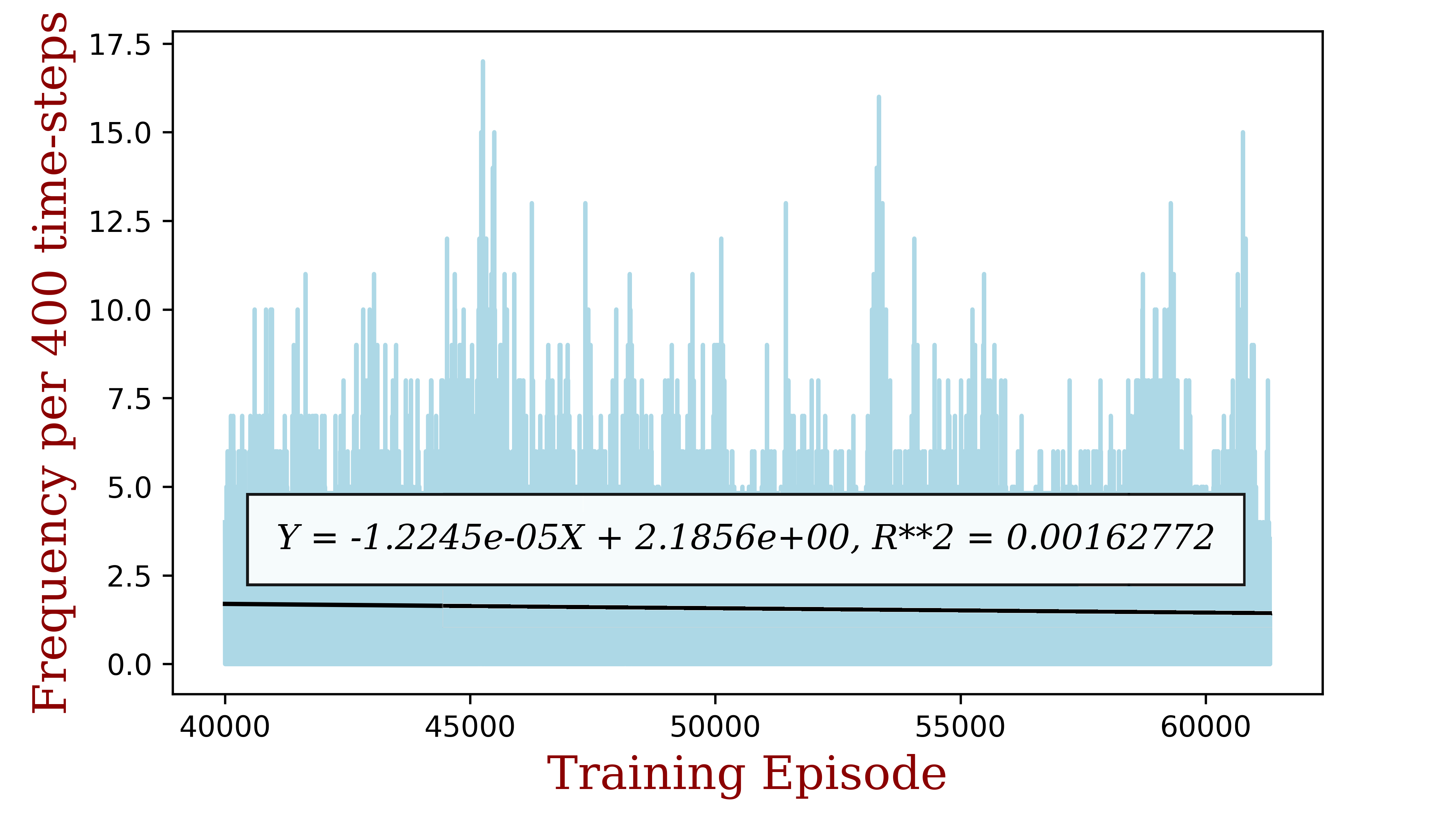}
	\caption{ Each line shows the number of times \enhancedAgent{}
          chose to increase the sending rate three consecutive times
          or more, during the training process. This time the agent
          performed only 1-2 violations on average per training episode, which translates to a frequency of about 0.34\%. The resulting linear regression line is $y = -0.000012245x + 2.1856$.}
    	\label{fig:constrained_aurora_training_3some_fequency_plot}
\end{figure}

In contrast, for \enhancedAgent{}, the average frequency of performing
a violation is about 0.34\%, as can be seen in
Fig.~\ref{fig:constrained_aurora_training_3some_fequency_plot}.

\noindent
\textbf{Summary.}  The results above demonstrate a highly significant
change in behavior between \originalAgent{} and \enhancedAgent{} when
it comes to the frequency of performing a violation: whereas
\originalAgent{} would perform a violation about 9-11\% of the time,
for \enhancedAgent{} this rate drops to 0.34\%. Both agents, however,
achieve a similar overall reward level, indicating that they both learned
an adequate policy with respect to the main goal's of the
system. While the enhanced agent took longer to converge to this
policy (as it had to learn additional constraints), these
results showcase the feasibility and potential of our approach.

\section{\uppercase{Related Work}}\label{relatedwork}
Several approaches have been proposed in recent years for using
hand-crafted software components to enhance the run-time functionality
and performance of DNNs, or to improve their training process. One
notable family of approaches calls for \emph{composing} DNNs and
hand-crafted components. 
See for example~\cite{shalev2017formal} where the decisions of an autonomous driving systems can be overridden by rules.
This composition can be parallel, where a DNN 
and hand-crafted code run side by side, each handling different tasks;
sequential, where a DNN's output feeds as input into hand-crafted
code, or vice-versa; or ensemble-based, where
DNNs and hand-crafted code attempt to solve the same problem and agree
on an output. 

Another notable family of approaches contains those that are
\emph{reflection-based}, where both hand-crafted code and DNN
gradually adjust themselves according to their past
performance~\cite{kang2017detection,milan2017online,ray2020mixed}.
The approach we propose here can be viewed as a bridge between 
composition-based 
and reflection-based approaches: a hand-crafted model is run alongside
a DRL agent, with the purpose of improving the latter's training
process.

Prior work has explored the potential synergies between SBM and DRL. In one attempt, 
Elyasaf et al.~\cite{elyasaf2019using} used DRL to fine-tune the execution strategy of an existing scenario-based model. Using a game of RoboSoccer as a case-study,  they demonstrated how the DRL agent could learn to guide a scenario-based player to more effectively grab the soccer ball. In a separate attempt, Katz~\cite{katz2020guarded,Ka21b} focused on using SBM models to \emph{guard} an existing
DRL agent --- that is, to override decisions made by the DRL agent that
violate the scenario-based model. The technique that we propose here is different from and complementary to both of these approaches: instead of using DRL to guide a hand-crafted model or using SBM to guard an existing DRL agent, we propose to use SBM to improve the DRL agent,
a-priori, so that it better abides by a scenario-based specification. 

\section{CONCLUSION AND FUTURE WORK}\label{conclusion}

Deep reinforcement learning is an excellent tool for addressing many
real-world problems; but it is lacking, in the sense that it does not
naturally lend itself to the integration of expert knowledge. Through
our proposed approach, we seek to bridge this gap, and allow the
integration of classical modeling techniques into the DRL training
loop. The resulting agents, as we have demonstrated, are more likely
to adhere to the policies, goals and restrictions 
defined by the domain experts. Apart from improving performance, this approach
increases the transparency and explainability of DRL agents, and can be regarded as
documenting and explaining these agents' behavior.

Turning to the future, we intend to apply the technique to more complex
case-studies, involving more intricate agents and more 
elaborate scenario-based specifications
--- and use them to also study the scalability of the approach.
Another angle we intend to pursue is to
enhance the DRL-SBM interface, either by changing the SBM semantics or
by defining an event-based protocol so that it can allow manipulating
the agent's reward function in more subtle ways.  Thus, the SBM
feedback will be able to distinguish  actions that are slightly
undesirable from those that are extremely undesirable. We also intend to
explore adding constructs for encouraging an agent to take desirable
actions, in addition to penalizing it for taking undesirable ones. As
discussed earlier, we are also interested in exploring generalizable
criteria for choosing penalty values that will be conducive to
learning the desired properties, while preserving the overall learning
of the task at hand.

Another angle for future research is to measure additional effects that
scenario-assisted training 
may have on DRL, such as accelerating the learning of properties that could, in principle, be
learned without such assistance. It is also interesting to see on a
variety of case studies if SBM-specified expert advice that is aimed
at improving system performance (as opposed to complying with new
requirements) indeed accomplishes such improvement over what the
system could learn on its own.

\vfill
\section*{\uppercase{Acknowledgements}}
The work of R.~Yerushalmi, G.~Amir, A.~Elyasaf and G.~Katz was partially supported by a grant from the Israeli Smart Transportation Research Center (ISTRC). 
The work of D. Harel, A.~Marron and R.~Yerushalmi was partially supported by a research grant from the Estate of Ursula Levy, the Midwest Electron Microscope Project, Dr. Elias and Esta Schwartz Instrumentation Fund in Memory of Uri and Atida Litauer, and the Crown Family Philanthropies.

\bibliographystyle{apalike}
{\small
\bibliography{sbpForRLTrainingBib}}

\end{document}